\documentclass[letterpaper]{article} 
\usepackage{aaai2026}  
\usepackage{times}  
\usepackage{helvet}  
\usepackage{courier}  
\usepackage[hyphens]{url}  
\usepackage{graphicx} 
\usepackage{natbib}  
\usepackage{caption} 
\usepackage{algorithm}
\usepackage{algorithmic}

\usepackage{newfloat}
\usepackage{listings}
\DeclareCaptionStyle{ruled}{labelfont=normalfont,labelsep=colon,strut=off} 
\floatstyle{ruled}
\newfloat{listing}{tb}{lst}{}
\floatname{listing}{Listing}
\usepackage{amsmath}
\usepackage{booktabs}
\usepackage{multirow}
\usepackage{hhline} 
\usepackage{makecell}

\title{NICE: Neural Implicit Craniofacial Model for Orthognathic Surgery Prediction}
\author{
    Jiawen Yang\textsuperscript{\rm 1}\equalcontrib, Yihui Cao\textsuperscript{\rm 1}\equalcontrib, Xuanyu Tian\textsuperscript{\rm 1}, Yuyao Zhang\textsuperscript{\rm 1}, Hongjiang Wei\textsuperscript{\rm 2}\thanks{Corresponding author.}
}
\affiliations{

    \textsuperscript{\rm 1} School of Information Science and Technology, ShanghaiTech University, Shanghai 201210, China\\
    \textsuperscript{\rm 2} School of Biomedical Engineering, Shanghai Jiao Tong University, Shanghai 200127, China\\
    yangjw2023@shanghaitech.edu.cn, caoyh2025@shanghaitech.edu.cn, tianxy@shanghaitech.edu.cn, zhangyy8@shanghaitech.edu.cn, hongjiang.wei@sjtu.edu.cn
}

\usepackage{bibentry}

\begin{document}

\maketitle

\begin{abstract}
Orthognathic surgery is a crucial intervention for correcting dentofacial skeletal deformities to enhance occlusal functionality and facial aesthetics. Accurate postoperative facial appearance prediction remains challenging due to the complex nonlinear interactions between skeletal movements and facial soft tissue. Existing biomechanical, parametric models and deep-learning approaches either lack computational efficiency or fail to fully capture these intricate interactions. 
To address these limitations, we propose Neural Implicit Craniofacial Model (NICE) which employs implicit neural representations for accurate anatomical reconstruction and surgical outcome prediction. NICE comprises a shape module, which employs region-specific implicit Signed Distance Function (SDF) decoders to reconstruct the facial surface, maxilla, and mandible, and a surgery module, which employs region-specific deformation decoders. These deformation decoders are driven by a shared surgical latent code to effectively model the complex, nonlinear biomechanical response of the facial surface to skeletal movements, incorporating anatomical prior knowledge. The deformation decoders output point-wise displacement fields, enabling precise modeling of surgical outcomes.
Extensive experiments demonstrate that NICE outperforms current state-of-the-art methods, notably improving prediction accuracy in critical facial regions such as lips and chin, while robustly preserving anatomical integrity. This work provides a clinically viable tool for enhanced surgical planning and patient consultation in orthognathic procedures.
\end{abstract}

\begin{links}
    \link{Code}{https://github.com/Bill-Yang-1/NICE}
\end{links}

\section{Introduction}
Orthognathic surgery is a principal clinical intervention for maxillofacial deformities, undertaken to correct dentofacial skeletal malocclusions and thereby improve both occlusal functionality and facial aesthetics~\cite{posnick2021orthognathic}. Within the surgical workflow, preoperative planning is especially critical: accurately forecasting postoperative soft-tissue appearance not only facilitates surgeon–patient communication and optimizes surgical strategy but also improves overall patient satisfaction~\cite{stokbro2014virtual}. Facial soft tissue responds to skeletal movement through highly complex, nonlinear couplings that remain only partially understood, making postoperative facial appearance prediction a formidable challenge.

Early solutions focused on biomechanical model notably the mass-spring models (MSMs), mass-tensor models (MTMs) and finite-element models (FEMs)~\cite{nadjmi2014quantitative,knoops2018novel,kim2019new}. An incremental FEM with realistic lip-sliding effects (FEM-RLSE) currently delivers the most precise perioral predictions, yet its accuracy depends on labour-intensive, patient-specific model construction, resulting in lengthy preparation times and high computational cost~\cite{kim2021novel}. These factors restrict its scalability and real-time applicability in routine clinical settings.

Three-dimensional parametric facial models have advanced facial geometry reconstruction by introducing structural constraints. Models such as Faces Learned with an Articulated Model and Expressions (FLAME)~\cite{li2017learning} and Neural Parametric Head Models (NPHM)~\cite{giebenhain2023learning} achieve efficient soft tissue fitting but these models only fit the soft tissue, omitting the mandible and maxilla. SCULPTOR~\cite{qiu2022sculptor} extends this paradigm by jointly modelling the skull and facial soft tissue with a skeleton-driven strategy that maintains anatomical consistency and enhances realism, showing promise for postoperative facial appearance prediction. Nevertheless, its reliance on a linear blend skinning (LBS) limits its ability to capture the complex nonlinear deformation between skeleton and soft tissue.

The rapid progress of deep learning offers a new route for postoperative facial appearance prediction. Data-driven networks~\cite{lampen2023spatiotemporal,ma2022simulation,ma2023bidirectional,fang2024correspondence,bao2024deep,huang2025maxillofacial} learn the intricate nonlinear mapping between skeletal movements and soft-tissue deformation from paired preoperative and postoperative data, markedly improving computational efficiency and automation.  Building on this paradigm, several studies~\cite{yang2025blending} have augmented the SCULPTOR framework with neural components, capturing subtle interactions between skeleton and soft tissue. Nevertheless, these approaches—like most current neural models—are supervised mainly by point-to-point distances. This neglect of explicit geometric relationships among neighbouring points and surface connectivity hampers continuity and normal-vector consistency. Moreover, the absence of anatomical and kinematic constraints can allow predicted skeletal meshes to penetrate the skin surface, compromising structural fidelity and clinical reliability.

\begin{figure*}[t]
\centering
\includegraphics[width=0.9\textwidth]{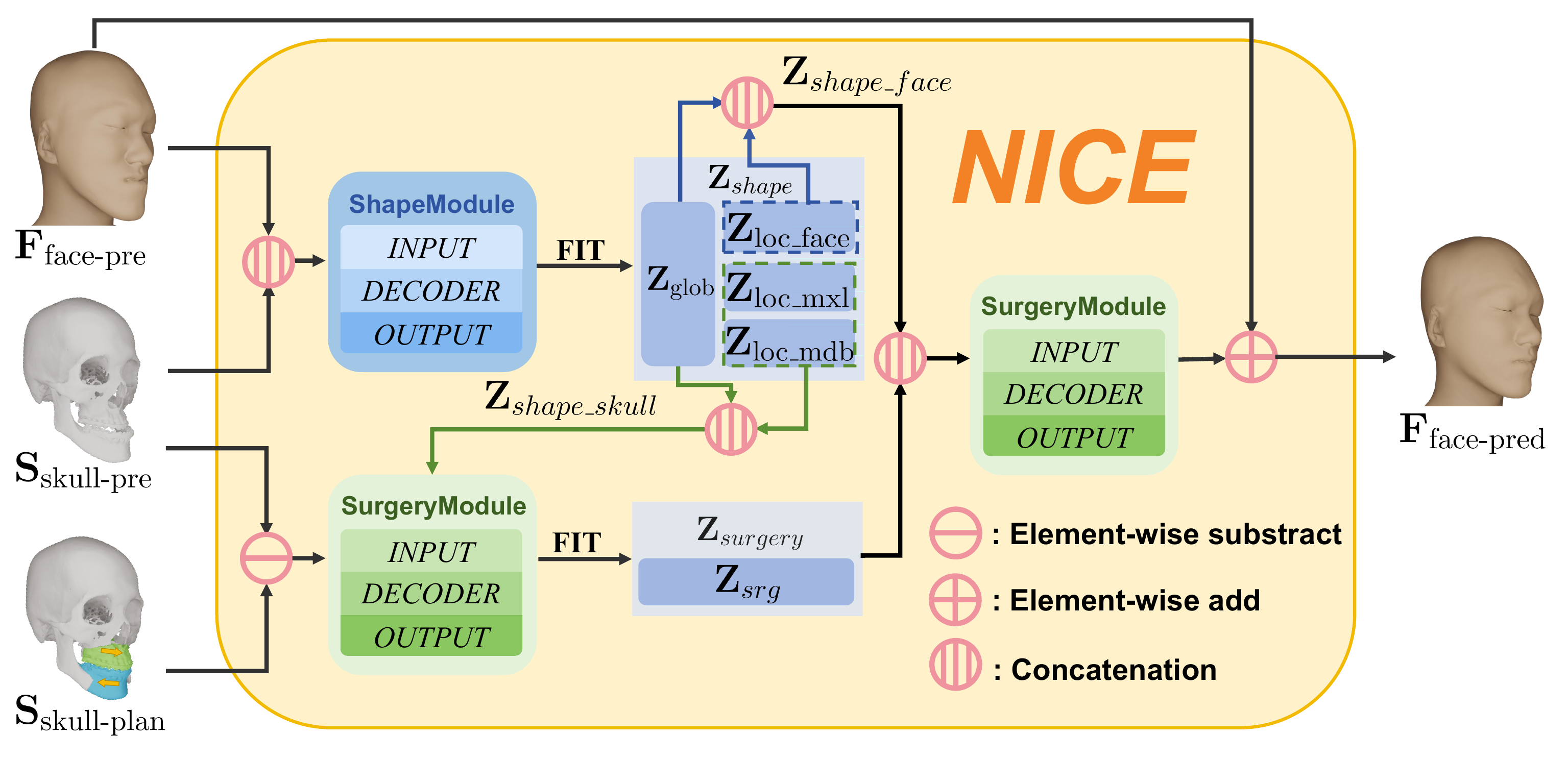} 
\caption{Pipeline of our neural implicit craniofacial model (NICE) for postoperative facial appearance prediction.
The shape module first reconstructs the preoperative facial surface $\mathbf{F}_\text{face-pre}$ and skull $\mathbf{S}_\text{skull-pre}$ from CT scans, generating the corresponding shape latent codes $\mathbf{Z}_{shape\_face}$ and $\mathbf{Z}_{shape\_skull}$. The surgery module then fits a shared surgery latent code $\mathbf{Z}_{srg}$ by matching the deformation from $\mathbf{S}_\text{skull-pre}$ to the surgically planned postoperative skull $\mathbf{S}_\text{skull-plan}$. Finally, $\mathbf{Z}_{srg}$, $\mathbf{Z}_{shape\_face}$ and $\mathbf{F}_\text{face-pre}$ are combined in the surgery module to predict the postoperative facial surface $\mathbf{F}_\text{face-pred}$.
}
\label{pipeline}
\end{figure*}

To address these limitations while simultaneously ensuring anatomical consistency and expressive nonlinear modeling, we propose Neural Implicit Craniofacial Model (NICE) for orthognathic surgery prediction. Our key contributions are summarized as follows: 
\begin{itemize}
    \item A shape module employing region-specific implicit SDF decoders (for face, maxilla, mandible) built around localized landmarks, enabling high-fidelity reconstruction of both bony structures and facial surface.
    \item A surgery module utilizing region-specific deformation decoders driven by a shared surgery latent code, effectively modeling the complex, nonlinear biomechanical response of facial surface to skeletal movements while leveraging anatomical priors.
    \item An integrated framework providing a streamlined workflow for accurate postoperative facial appearance prediction. Extensive experiments demonstrate that NICE  outperforms state-of-the-art methods in prediction fidelity, particularly in critical regions like lips and chin, while robustly maintaining anatomical consistency.
\end{itemize}

\section{Related Work}
\subsection{Conventional Biomechanical Models}

Conventional biomechanical approaches to postoperative facial appearance prediction fall into three main categories: the mass–spring models (MSMs), mass–tensor models (MTMs) and finite element methods (FEMs)~\cite{nadjmi2014quantitative,knoops2018novel,kim2019new}. MSM models soft tissue as point masses connected by springs governed by Hooke’s law. MTM extends MSM by assigning each mass a local stiffness tensor, enabling spatially localized, direction‑dependent rigidity and hence anisotropic stress–strain behavior. FEM discretizes maxillofacial skeleton and soft tissue into three‑dimensional elements and solves linear‑elastic or nonlinear equations. The incremental FEM with realistic lip‑sliding effects (FEM‑RLSE) imposes explicit lip‑sliding constraints and staged skeletal increments, which is widely regarded as the most precise simulator of postoperative facial change~\cite{kim2021novel}. Despite their accuracy, biomechanical models entail patient‑specific meshing and heavy computation. End-to-end processing times frequently reach several hours per case, severely limiting scalability and real-time applicability in routine clinical practice.

\subsection{Head Parametric Models}
Parametric models encode craniofacial variation in a differentiable, low-dimensional latent space, offering an attractive compromise between computational efficiency and controllability. FLAME~\cite{li2017learning} learns identity and expression shape spaces via PCA and employs LBS to couple them. NPHM~\cite{giebenhain2023learning} represents facial geometry as a continuous SDF, achieving resolution‑independent detail. However, most parametric models deform only the soft tissue surface, omitting skeleton‑driven mechanics essential to orthognathic planning.
To address this gap, SCULPTOR~\cite{qiu2022sculptor} extends the human face model FLAME~\cite{li2017learning} to a physically precise parametric human skeleton model with the canonical pose.  It introduces a trait component to model physiologically plausible variations in the internal skeletal structure of the face, as well as the associated shape changes that occur due to modifications in the skeleton. The trait component improves the model’s ability to generate more characteristic faces with a physiologically correct constraint.
Nevertheless, the linear additive nature of LBS limits its ability to capture nonlinear relationships. This limitation constrains SCULPTOR and similar parametric models from fully representing the intricate nonlinear interactions between craniofacial skeleton and soft tissue. Thus they fall short of the precision demanded by clinical postoperative facial appearance prediction.

\subsection{Deep Learning Models}
Recent deep learning advances have led to more efficient methods for predicting postoperative facial appearance. PointNet~\cite{qi2017pointnet} introduced end-to-end geometric-feature learning on unordered point clouds, eliminating the need for meshes or voxelized representations. Inspired by this, subsequent studies represent skull–face morphology as three-dimensional point clouds and train networks to learn per-point displacement fields that map preoperative to postoperative anatomy.
A representative example is ACMT‑Net~\cite{fang2024correspondence} which introduces an Attentive Correspondence assisted Movement Transformation network to correlate skeletal displacements with facial soft-tissue changes, achieving higher computational efficiency than FEM-based methods while maintaining accuracy. Another approach~\cite{yang2025blending} integrates neural networks into the SCULPTOR parametric framework, modeling nonlinear skeleton-to-soft-tissue deformation and improving reconstruction accuracy.
However, most networks use Chamfer distance or point-to-point losses without explicit constraints on local geometry and mesh connectivity. This can lead to discontinuities, inconsistent normals, and physiologically implausible deformations, limiting the reliability needed for clinical use.

\section{Method}
\subsection{Overview}
Our neural implicit craniofacial model primarily consist of two modules: the shape module and the surgery module. The shape module represents the anatomical geometry of the facial outer surface and underlying skull within the canonical space. The core concept of the surgery module centers on biomechanical deformations, which constrain the plausible transformation space for both the facial outer surface and skull during surgical procedures. 
In addition, we present a postoperative facial appearance prediction framework as shown in Fig.~\ref{pipeline}.

\subsection{Shape Module}

\begin{figure}[t]
\centering
\includegraphics[width=0.98\columnwidth]{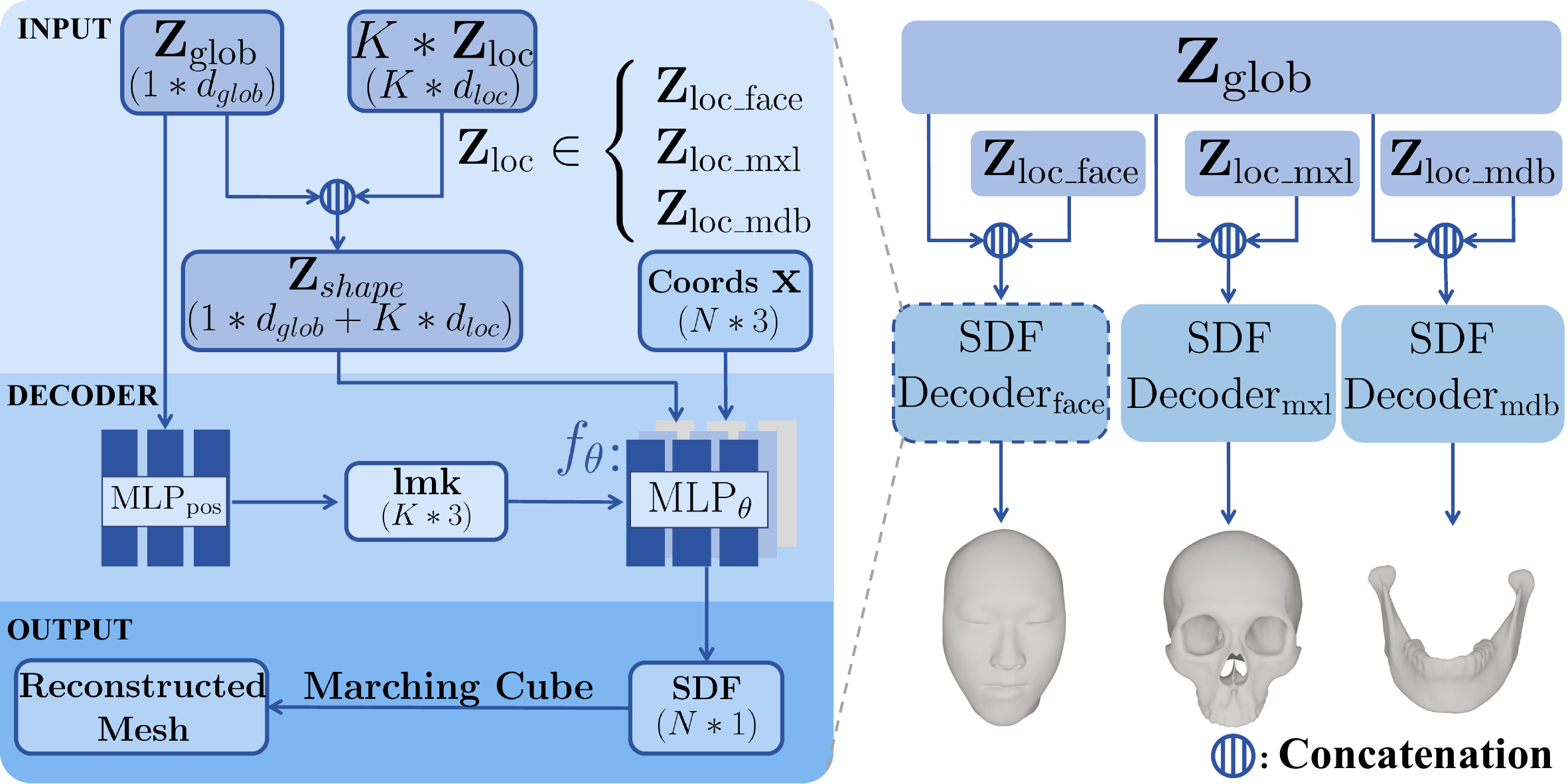} 
\caption{Architecture of the shape module in NICE. Region-specific SDF decoders reconstruct facial, maxillary, and mandibular geometry from spatial coordinates and shape latent codes, using landmark-centered MLP ensembles.}
\label{shape}
\end{figure}

We implicitly represent person-specific head geometry using canonical-space signed distance functions (SDFs).
As shown in Fig.~\ref{shape}, our shape module employs three region-specific SDF decoders for the facial surface, maxilla, and mandible.  
For each SDF decoder, we decompose it into an ensemble of several smaller MLPs, each constructed around $K$ specific landmarks $\mathbf{lmk}\in\mathbf{R}^{K\times3}$.
We define the integrated SDF decoder through the following equation:
\begin{equation}
\mathcal{F}_{shape}(\mathbf{x}, \mathbf{Z}_{shape}) = \sum_{k=0}^{K} w_k(\mathbf{x}, \mathbf{lmk}^k) f_{\theta}^k(\mathbf{x}, [\mathbf{Z}_\text{glob}, \mathbf{Z}_\text{loc}^k]).
\label{Zshape decoder}
\end{equation}
Specifically, the inputs to each SDF decoder are: (1) query spatial coordinates $\mathbf{x}\in\mathbf{R}^3$, and (2) region-specific shape latent code $\mathbf{Z}_{shape}$ which represents the geometric characteristics of its anatomical region: face, maxilla or mandible.


$\mathbf{Z}_{shape}$ consists of a globally shared latent vector $\mathbf{Z}_\text{glob} \in \mathbf{R}^{d_{glob}}$ and $K$ region-specific local latent vectors $\mathbf{Z}_\text{loc}^k \in \mathbf{R}^{d_{loc}}$. The same $\mathbf{Z}_\text{glob}$ is used across the facial, maxillary, and mandibular SDF decoders, while each $\mathbf{Z}_\text{loc}^k$ corresponds to the $k$-th latent vector from $K$ localized MLPs, associated with the facial surface (denoted as $\mathbf{Z}_\text{loc\_face}$), maxilla ($\mathbf{Z}_\text{loc\_mxl}$), or mandible ($\mathbf{Z}_\text{loc\_mdb}$), respectively.
To enhance $\mathbf{Z}_\text{glob}$'s representation capacity and constrain the semantic scope of local MLPs, we employ a compact $\text{MLP}_{\text{pos}}$ to predict these $K$ landmark positions $\mathbf{lmk}$ from $\mathbf{Z}_\text{glob}$.
The weights $w_k(\mathbf{x}, \mathbf{lmk}^k)$ in Eq. (\ref{Zshape decoder}) are computed using Gaussian kernels, as commonly done in \cite{genova2020local,zheng2022structured}.
The $f_{\theta}^k$ in Eq. (\ref{Zshape decoder}) denotes the $k$-th small $\text{MLP}_{\theta}$, defined as follows:
\begin{equation}
f_{\theta}^k(\mathbf{x}, [\mathbf{Z}_\text{glob}, \mathbf{Z}_\text{loc}^k]) 
    = \text{MLP}_{\theta}^k(\mathbf{x}-\mathbf{lmk}^k, [\mathbf{Z}_\text{glob}, \mathbf{Z}_\text{loc}^k]),
\label{small mlp}
\end{equation}
where $[\cdot]$ represents the concatenation operation.
Specifically, to handle SDF values in regions distant from all landmarks in $\mathbf{lmk}$, we set $\mathbf{lmk}^0 = 0 \in \mathbf{R}^3$ and introduce the following definition:
\begin{equation}
f_{\theta}^0(\mathbf{x}, [\mathbf{Z}_\text{glob}, \mathbf{Z}_\text{loc}^0]) 
    = \text{MLP}_{\theta}^0(\mathbf{x}, [\mathbf{Z}_\text{glob}, \mathbf{Z}_\text{loc}^0]).
\label{f theata 0}
\end{equation}
Since the SDF decoder outputs SDF values at query coordinates $\mathbf{x}$, we can then extract the region-specific mesh through spatial sampling via the marching cubes algorithm.

\subsection{Surgery Module}
\begin{figure}[t]
\centering
\includegraphics[width=0.98\columnwidth]{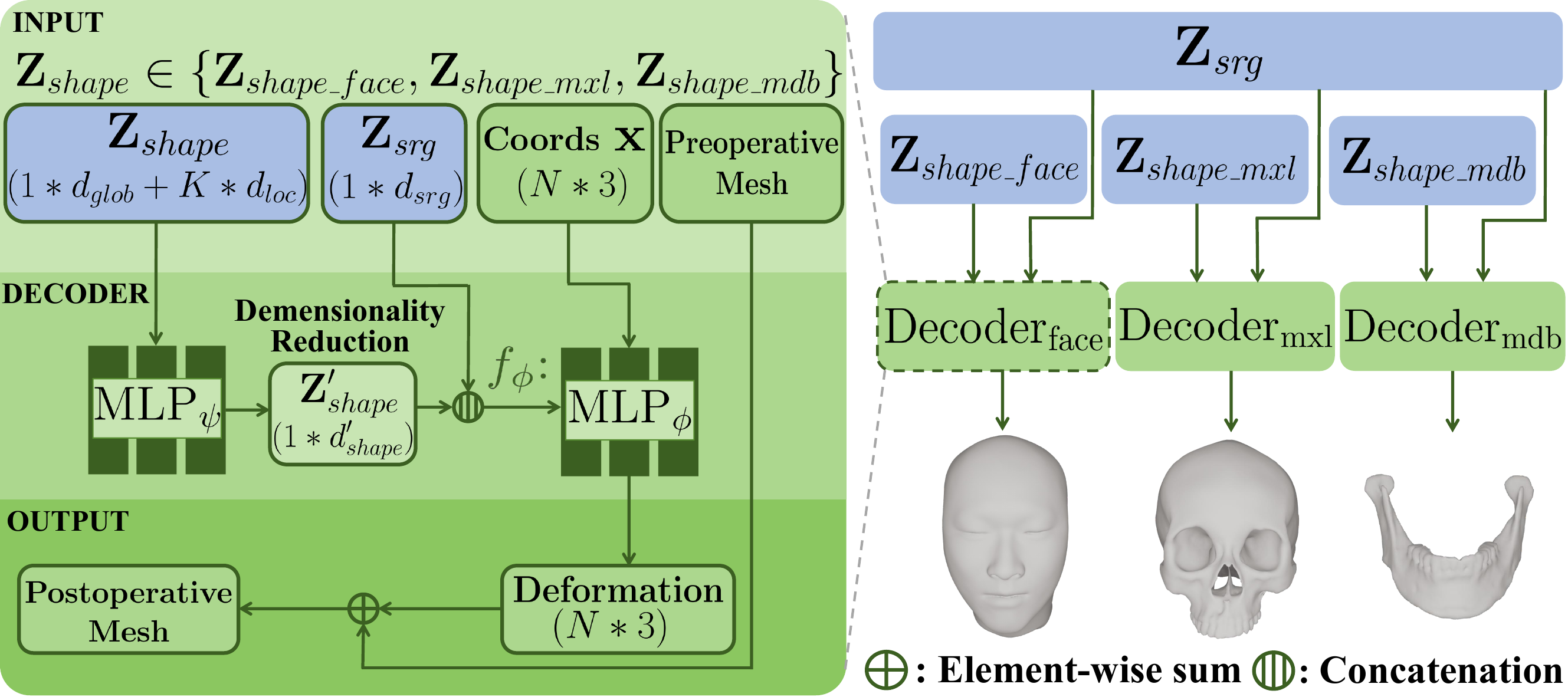} 
\caption{
Architecture of the surgery module in NICE. Region-specific decoders predict deformation fields from spatial coordinates, a shared surgery latent code, and shape latent priors, enabling personalized postoperative mesh generation.}
\label{surgery}
\end{figure}
The surgery module models the transformation of the facial surface and skull from preoperative to postoperative states using deformation fields. As shown in Fig.~\ref{surgery}, it incorporates three region-specific decoders for the facial surface, maxilla, and mandible, similar to the shape module.
We define this decoder through the following equation:
\begin{equation}
\mathcal{F}_{srg}(\mathbf{x}, \mathbf{Z}_{srg}, \mathbf{Z}_{shape}) = f_{\phi}(\mathbf{x}, [\mathbf{Z}_{srg}, \text{MLP}_{\psi}(\mathbf{Z}_\text{glob}, \mathbf{Z}_\text{loc})]).
\label{Zsurgery decoder}
\end{equation}
Here, $f_{\phi}$ denotes the main $\text{MLP}_{\phi}$ that predicts the deformation field based on the concatenated latent representation. 
Specifically, the inputs to each decoder are: (1) query spatial coordinates $\mathbf{x}$, (2) surgery latent code $\mathbf{Z}_{srg}\in\mathbf{R}^{d_{srg}}$ simultaneously constraining the plausible transformation space for facial surface and skull structures during surgical procedures, and (3) shape latent code $\mathbf{Z}_{shape}$, serving as anatomical prior, obtained from the shape module by fitting the geometry of the corresponding region.
The surgery latent code $\mathbf{Z}_{srg}$ is globally shared across the three decoders, i.e., the same $\mathbf{Z}_{srg}$ is used for the facial surface, maxilla, and mandible deformation.
The $\text{MLP}_{\psi}$ in Eq. (\ref{Zsurgery decoder}) performs dimensionality reduction on $\mathbf{Z}_{shape}$ to reduce computational cost.
Since this decoder outputs the displacement from preoperative to postoperative states at query coordinates $\mathbf{x}$, we can obtain the postoperative mesh by applying this displacement to the preoperative person-specific head mesh.

\subsection{Postoperative Facial Appearance Prediction}
The pipeline of our postoperative facial appearance prediction framework is shown in Fig.~\ref{pipeline}. According to Eq. (\ref{Zshape decoder}), we first employ the shape module to reconstruct the patient's preoperative facial surface $\mathbf{F}_\text{face-pre}$, and skull structures maxilla $\mathbf{S}_\text{mxl-pre}$ and mandible $\mathbf{S}_\text{mdb-pre}$ obtained from CT scan data:
\begin{equation}
\begin{aligned}
\mathbf{F}_\text{face-pre} &= \mathcal{F}_{shape}(\mathbf{x}_{face}, \mathbf{Z}_{shape\_face})  \\
\mathbf{S}_\text{mxl-pre} &= \mathcal{F}_{shape}(\mathbf{x}_{mxl}, \mathbf{Z}_{shape\_mxl}) \\
\mathbf{S}_\text{mdb-pre} &= \mathcal{F}_{shape}(\mathbf{x}_{mdb}, \mathbf{Z}_{shape\_mdb}) 
\end{aligned}
\end{equation}
This yields patient-specific shape latent codes $\mathbf{Z}_{shape\_face}$, $\mathbf{Z}_{shape\_mxl}$, and $\mathbf{Z}_{shape\_mdb}$, encoding the unique geometry of each anatomical region. For brevity, we denote the combined skull structures and their corresponding shape latent codes as $\mathbf{S}_\text{skull-pre} = \{\mathbf{S}_\text{mxl-pre}, \mathbf{S}_\text{mdb-pre}\}$ and $\mathbf{Z}_{shape\_skull} = \{\mathbf{Z}_{shape\_mxl}, \mathbf{Z}_{shape\_mdb}\}$, respectively.

Based on Eq. (\ref{Zsurgery decoder}), we use the surgery module to fit the shared surgery latent code $\mathbf{Z}_{srg}$ to represent the planned bony movements. This is achieved by matching the deformation between the preoperative skull $\mathbf{S}_\text{skull-pre}$ and the surgically planned postoperative skull $\mathbf{S}_\text{skull-plan}$:
\begin{equation}
\mathcal{F}_{srg}(\mathbf{x}_{skull}, \mathbf{Z}_{srg}, \mathbf{Z}_{shape\_skull}) = \mathbf{S}_\text{skull-plan} - \mathbf{S}_\text{skull-pre}.
\end{equation}

Critically, given that the $\mathbf{Z}_{srg}$ simultaneously constrains the plausible transformation space for both the facial surface and skull during surgery,  we combine it with the facial shape latent code $\mathbf{Z}_{shape\_face}$ to deform the preoperative facial surface $\mathbf{F}_\text{face-pre}$ and predict the postoperative facial outcome:
\begin{equation}
\mathbf{F}_\text{face-pred} = \mathbf{F}_\text{face-pre} + \mathcal{F}_{srg}(\mathbf{x}_{face}, \mathbf{Z}_{srg}, \mathbf{Z}_{shape\_face}).
\end{equation}
The resulting $\mathbf{F}_\text{face-pred}$ is our framework's prediction of the patient's postoperative facial surface morphology.

\subsection{Losses}
\subsubsection{Shape Module} To supervise the learning of the shape latent space for facial surface (denoted as face), maxilla (mxl), and mandible (mdb), we design a composite loss function that enforces geometric fidelity, regularizes the latent space, and constrains anatomical landmarks.
We jointly train the shape latent space by minimizing the overall shape loss as follows:
\begin{equation}
\mathcal{L}_{shape} =  \lambda_{\text{glob}}\mathcal{L}_{\text{glob}} + \sum_{r \in \{\text{face}, \text{mxl}, \text{mdb}\}} \mathcal{L}_{\text{SDF}}^r +\lambda_{\text{lmk}}^r\mathcal{L}_{\text{lmk}}^r,
\label{shape loss}
\end{equation}
where $\mathcal{L}_{\text{glob}}$ is $l2$ regularization term to constrain the norm of the global latent vector $\mathbf{Z}_\text{glob}$ and $\mathcal{L}_{\text{lmk}}^r$ denotes the landmark regression loss for region $r \in \{\text{face}, \text{mxl}, \text{mdb}\}$, penalizing the deviation between predicted and ground truth landmarks. The region-specific SDF loss term $\mathcal{L}_{\text{SDF}}^r$ is defined as:
\begin{equation}
\begin{aligned}
\mathcal{L}_{\text{SDF}}^r &= 
\lambda_{surf}^r \left|\mathcal{F}_{shape}(\mathbf{x}_r) \right| \\
& + \lambda_{surf}^r \left\| \nabla\mathcal{F}_{shape}(\mathbf{x}_r) 
- \mathbf{n}(\mathbf{x}_r) \right\|_2 \\
& + \lambda_{eik}^r\left[ \left| \left\| \nabla\mathcal{F}_{shape}(\mathbf{x}_r)  \right\|_2 - 1 \right| \right] \\
& + \lambda_{far} \exp\left( -\alpha \cdot \left| \mathcal{F}_{shape}(\mathbf{x}_r)  \right| \right).
\end{aligned}
\label{sdf loss}
\end{equation}
The first term encourages the predicted SDF values at or near the surface to be close to zero.
The second term enforces consistency between the gradients of the predicted SDF and the ground truth surface normals $\mathbf{n}(\mathbf{x}_r)$.
The third term imposes the Eikonal constraint by regularizing the gradient magnitude to be close to 1, which is essential for SDF correctness.
The final term penalizes near-zero SDF predictions at far-field locations by applying an exponential penalty, thereby promoting well-separated SDF values away from the surface.

\subsubsection{Surgery Module} 
To effectively supervise the learning of a plausible deformation latent space that captures the complex biomechanical changes during orthognathic surgery, we design a loss function comprising correspondence consistency and deformation regularization for each anatomical region. We jointly optimize the surgery deformation latent space by minimizing the following composite loss:
\begin{equation}
\mathcal{L}_{surgery} = \lambda_{srg} \mathcal{L}_{srg} + \sum_{r \in \{\text{face}, \text{mxl}, \text{mdb}\}} \left|\left| \mathcal{F}_{srg}(\mathbf{x}_r) - \Delta(\mathbf{x}_r)\right|\right|_2,
\end{equation}
where $\mathcal{L}_{srg}$ is the latent regularization term that penalizes the $l_2$ norm of the surgery latent code $\mathbf{Z}_{srg}$ to avoid overfitting and ensure compactness of the latent space. $\Delta(\mathbf{x}_r)$ is defined as: $\Delta(\mathbf{x}_r) = \mathbf{P}_{\text{gt-post}}^r - \mathbf{P}_{\text{gt-pre}}^r,$ which represents the ground truth points displacement from preoperative to postoperative states for points $\mathbf{x}_r$.

\begin{figure}[!t]
  \centering
  \includegraphics[width=0.45\linewidth]{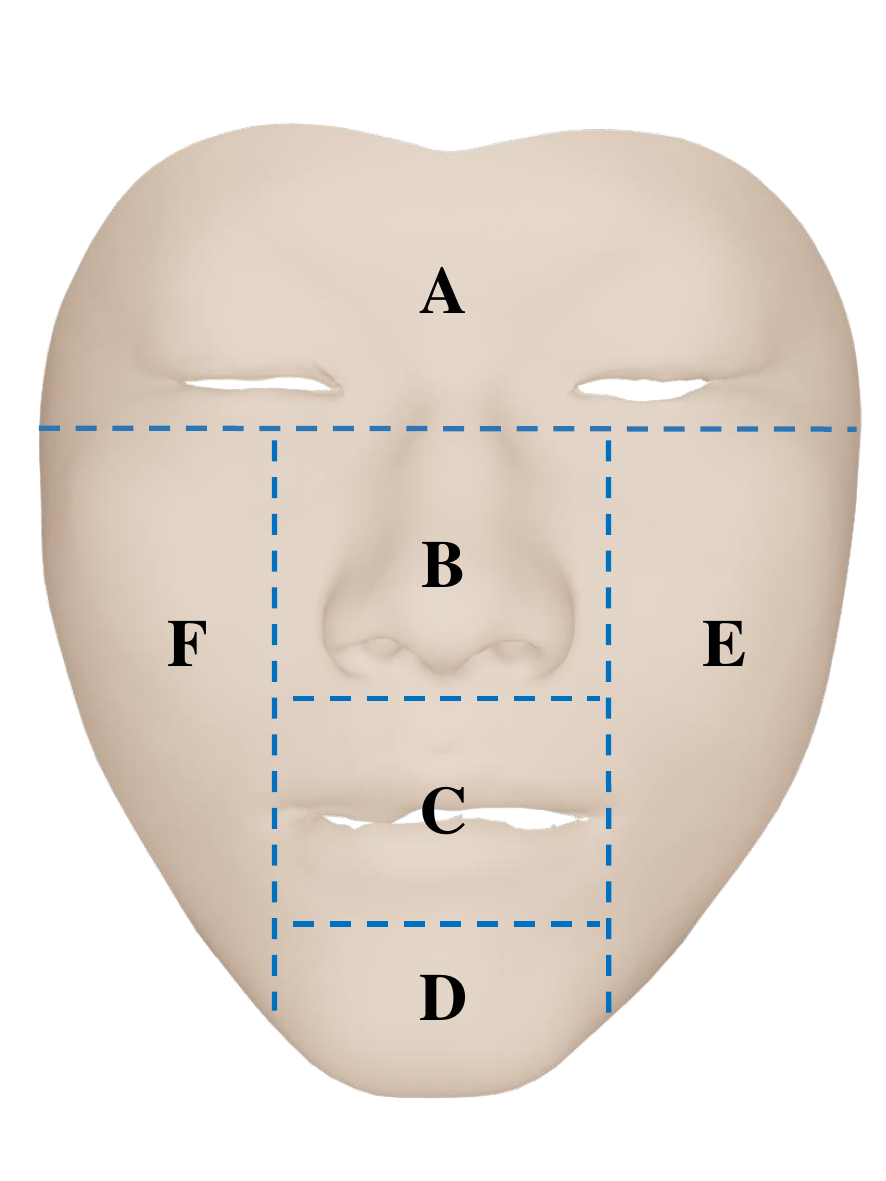}
  \caption{Illustration of facial sub-region division for quantitative evaluation.}
  \label{fig:different_regions}
\end{figure}

\section{Experiments and Results}
\subsection{Overview}
In this section, we conduct comprehensive experiments to evaluate the performance of our neural implicit craniofacial model. Specifically, we separately evaluate: (1) the shape module for anatomical geometry reconstruction, (2) the surgery module for biomechanical deformation modeling, and (3) the postoperative facial appearance prediction. 
\subsection{Experimental Settings}
\subsubsection{Dataset}
We accessed archived CT scan data from the clinical archives of the Department of Oral $\&$ Craniomaxillofacial Surgery at a collaborating hospital.
A total of 76 pairs of pre and post-surgery head CT images, along with multi-view facial scans, were collected, with 60 pairs used for training, 8 for testing and 8 for validation. 

\begin{table*}[t]
  \centering
  \setlength{\tabcolsep}{0.85mm}
  \begin{tabular}{c|cccc|cccc}
  
  \Xhline{0.5pt} 
    \hhline{|-|----|----|}
    \multirow{2}{*}{\textbf{Metric}} &
      \multicolumn{4}{c|}{\textbf{Shape module}} &
      \multicolumn{4}{c}{\textbf{Surgery module}} \\[-0.2ex] 
    \cline{2-9}
      & \textbf{Method} & \textbf{Face} & \textbf{Maxilla} & \textbf{Mandible} &
        \textbf{Method} & \textbf{Face} & \textbf{Maxilla} & \textbf{Mandible} \\
    \hhline{|-|----|----|}

    \multirow{3}{*}{\textbf{P2PL}$\downarrow$} &
      SCULPTOR-S                & 1.69$\pm$0.19 & 1.48$\pm$0.26 & 1.72$\pm$0.49 &
      SCULPTOR-T                & 1.14$\pm$0.16 & 1.12$\pm$0.12 & 1.09$\pm$0.18 \\
    & Ours(single decoder)      & \underline{0.62$\pm$0.16} & \underline{0.44$\pm$0.05} & \underline{0.54$\pm$0.21} &
      Ours(global $\mathrm{MLP}_{\psi})$   & \underline{0.76$\pm$0.17} & \underline{0.48$\pm$0.07} &  \textbf{0.42$\pm$0.08} \\
    & Ours(multi decoders)      &  \textbf{0.52$\pm$0.10} &  \textbf{0.42$\pm$0.05} &  \textbf{0.51$\pm$0.17} &
      Ours(separate $\mathrm{MLP}_{\psi})$ &  \textbf{0.73$\pm$0.17} &  \textbf{0.47$\pm$0.04} & \underline{0.44$\pm$0.09} \\
    \hhline{|-|----|----|}

    \multirow{3}{*}{\textbf{CD}$\downarrow$} &
      SCULPTOR-S                & 4.73$\pm$0.34 & 4.00$\pm$0.42 & 3.88$\pm$0.86 &
      SCULPTOR-T                & 3.69$\pm$0.28 & 3.19$\pm$0.24 & 2.73$\pm$0.28 \\
    & Ours(single decoder)      & \underline{3.17$\pm$0.20} & \underline{1.99$\pm$0.08} & \underline{1.71$\pm$0.25} &
      Ours(global $\mathrm{MLP}_{\psi})$   & \underline{2.27$\pm$0.44} & \underline{1.74$\pm$0.15} &  \textbf{1.29$\pm$0.20} \\
    & Ours(multi decoders)      &  \textbf{3.02$\pm$0.14} &  \textbf{1.93$\pm$0.07} &  \textbf{1.67$\pm$0.19} &
      Ours(separate $\mathrm{MLP}_{\psi})$ &  \textbf{2.25$\pm$0.38} &  \textbf{1.67$\pm$0.10} & \underline{1.35$\pm$0.19} \\
    \hhline{|-|----|----|}
    \Xhline{0.5pt} 
  \end{tabular}
  \caption{
Quantitative comparison of the shape and surgery modules, including ablation studies conducted against SCULPTOR-S and SCULPTOR-T. Evaluation is performed on the face, maxilla, and mandible, measured in millimeters (mm). Lower values indicate better performance. The best and second-best performances are shown in \textbf{bold} and \underline{underline}, respectively.}

  \label{tab:shape_surgery_joint}
\end{table*}

\subsubsection{Data Preprocessing}
To standardize coordinate systems across pre- and post-surgery CT scans and ensure reliable spatial correspondences for our surgery module, we performed a data registration based on predefined template face and skull models, following the approach in SCULPTOR~\cite{qiu2022sculptor}. The process includes an initial rigid alignment, followed by non-rigid deformation using embedded techniques to optimize vertex correspondence and anatomical landmarks. This results in meshes with consistent point and face counts, preserving topological consistency and integrity. Additional details are provided in the supplementary materials.

\subsubsection{Baselines and Metrics}

We adopt the parametric model SCULPTOR \cite{qiu2022sculptor}—which jointly represents facial surface and bony structures—as the benchmark for evaluating our shape and surgery modules. 
For assessment of postoperative facial appearance prediction, we conduct comparative experiments with recent state-of-the-art methods in craniofacial surgery prediction including FSC-Net \cite{ma2022simulation}, ACMT-Net \cite{fang2024correspondence}, and BPMNR \cite{yang2025blending}. We reproduce them and use the same implementation details as mentioned in their papers. 

For comprehensive comparisons, we use Point-to-Plane Distance (P2PL) and Chamfer Distance (CD) \cite{liu2020morphing}, both measured in millimeters (mm). Additionally, we provide a detailed regional analysis of postoperative facial  appearance prediction accuracy, with Fig.~\ref{fig:different_regions} showing the divisions: (A) forehead, (B) nose, (C) lips, (D) chin, (E) left cheek and (F) right cheek.

\begin{figure*}[t]
\centering
\includegraphics[width=1\textwidth]{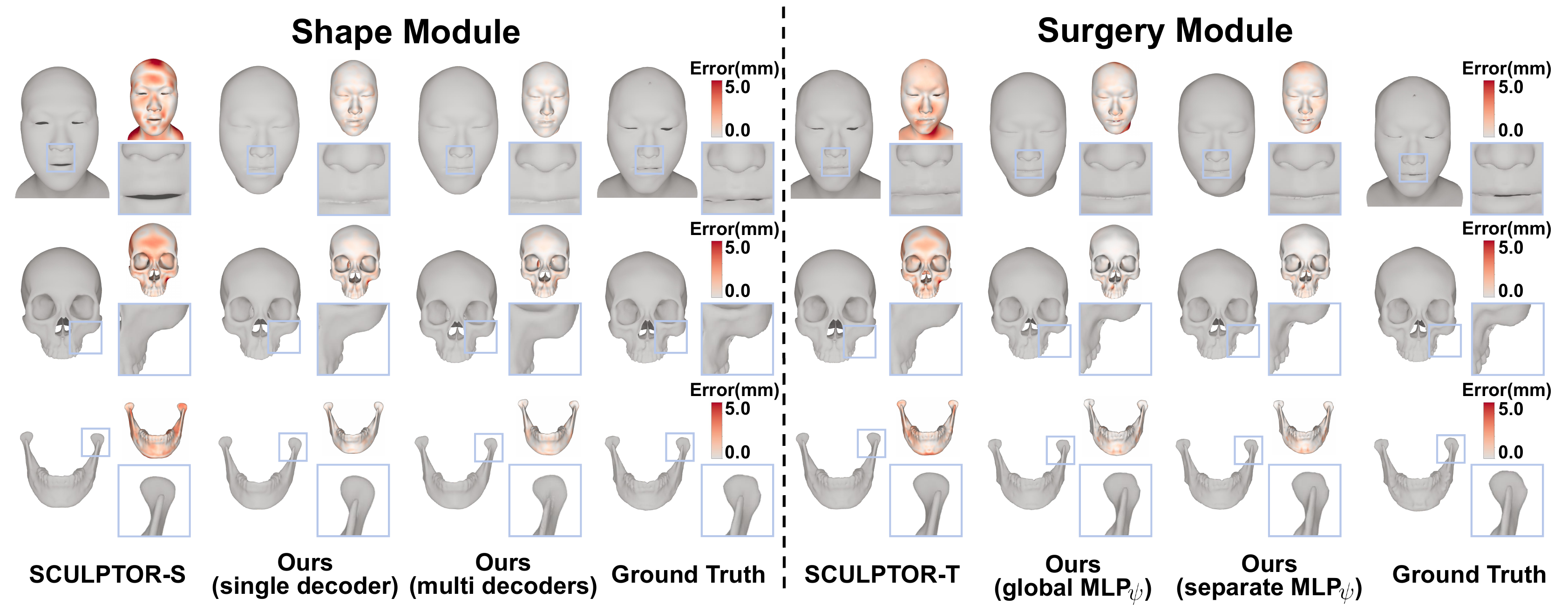} 
\caption{
Qualitative performance comparison of the shape and surgery modules, including ablation studies conducted against SCULPTOR-S and SCULPTOR-T. Each row corresponds to a different anatomical region: facial surface (top), maxilla (middle), and mandible (bottom). In each column, the top-right subfigure shows the P2PL error map compared with ground truth (value range: 0-5 millimeters), while the bottom-right subfigure displays magnified views of specific anatomical structures.}

\label{eval of shape and module}
\end{figure*}

\subsubsection{Implementation Details}
We use 39, 43, and 16 anatomical landmarks for the facial surface, maxilla, and mandible, respectively. The global latent code $\mathbf{Z}_\text{glob}$ has a dimension of 64. The local latent codes $\mathbf{Z}_\text{loc}$ for the facial surface, maxilla, and mandible are 32, 64, and 32-dimensional, respectively. The surgery latent code $\mathbf{Z}_{srg}$ is 512-dimensional. The local small MLP in Eq. (\ref{small mlp}) follows the DeepSDF architecture\cite{park2019deepsdf} with 4 hidden layers, each containing 200 hidden units.
Due to space constraints, additional network architectures and parameter settings are provided in the released code and the supplementary materials.
The shape module is trained for 30,000 epochs with initial learning rates of $5 \times 10^{-4}$ for network parameters and $1 \times 10^{-3}$ for shape latent codes. Both are halved every 3,000 epochs. The surgery module is trained for 9,000 epochs with the same initial learning rates, decayed every 600 epochs. Optimization is performed using the Adam optimizer~\cite{kingma2014adam}. All training is conducted on a single NVIDIA RTX 4090 GPU (24 GB).

\subsection{Evaluation of Shape and Surgery Modules}

\begin{figure}[!t]
\centering
\includegraphics[width=0.98\columnwidth]{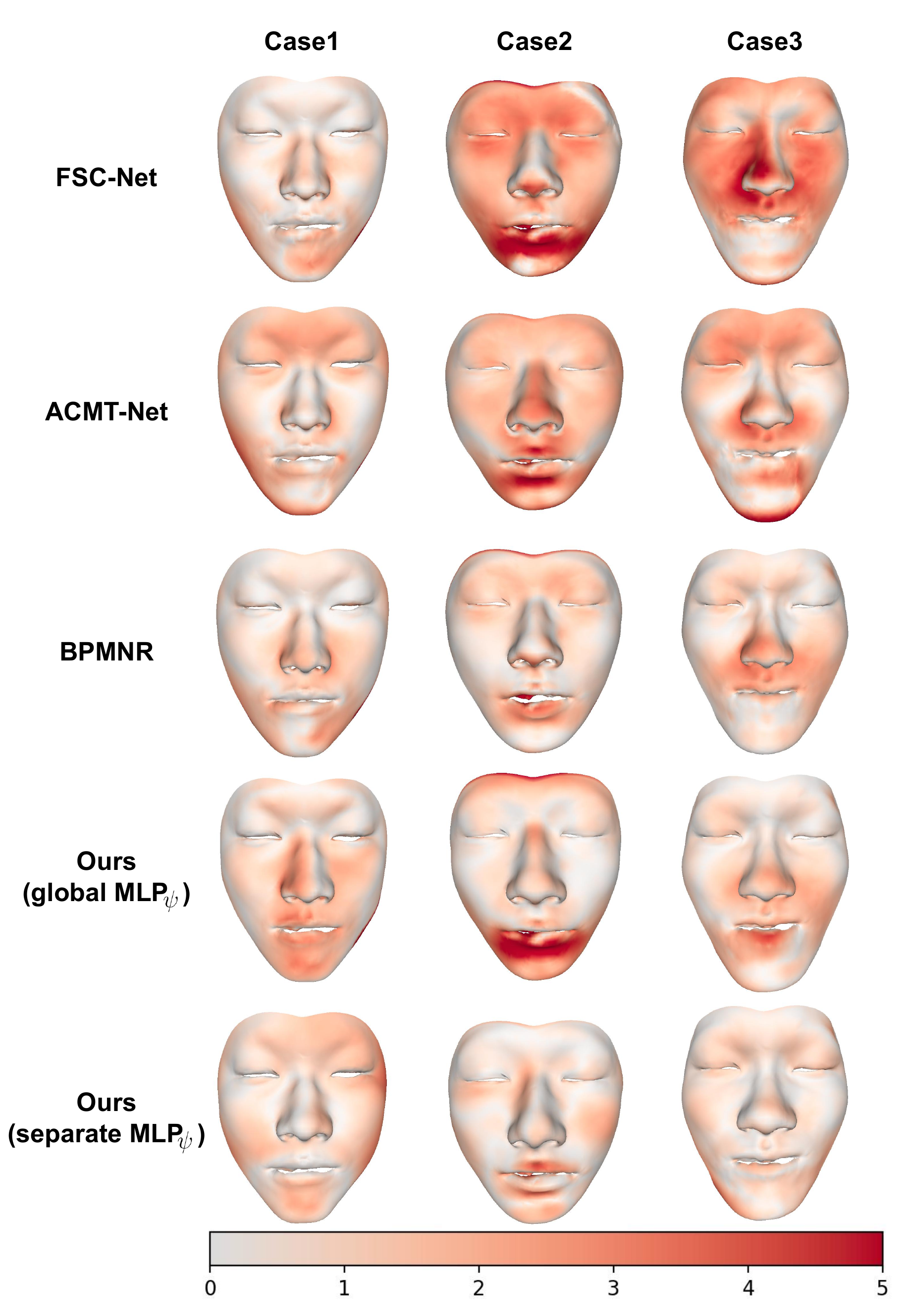} 
\caption{Qualitative P2PL error maps in comparison with FSC-Net, ACMT-Net, and BPMNR, including ablation studies on postoperative facial appearance prediction accuracy (value range: 0-5 millimeters).}
\label{surgery prediction}
\end{figure}

We employ the parametric model SCULPTOR as a benchmark to evaluate the capabilities of both our shape and surgery modules. 
Specifically, for shape module evaluation, we utilize SCULPTOR's shape component (denoted as SCULPTOR-S) for anatomical reconstruction, whereas for surgery module evaluation, we employ its trait component (denoted as SCULPTOR-T) to model surgical deformations.

Table~\ref{tab:shape_surgery_joint} summarizes the geometry reconstruction and deformation modeling accuracy of our shape and surgery modules, respectively. Compared to SCULPTOR-S and SCULPTOR-T, our method significantly reduces both P2PL and CD errors across all regions.
Fig.~\ref{eval of shape and module} illustrates visual comparisons of geometry reconstruction and surgery deformation against the ground truth. Our method produces smoother and more anatomically consistent surfaces, particularly in complex regions such as the lips, zygomatic bone, and temporomandibular joint (highlighted in the figure). The surgery module also better captures nonlinear deformations, yielding postoperative facial appearance predictions more closely aligned with the ground truth.
Both quantitative and qualitative results demonstrate that our method significantly enhances anatomical reconstruction and surgical deformation modeling accuracy, laying a strong foundation for reliable postoperative facial appearance prediction.

\subsection{Evaluation of Postoperative Facial Appearance Prediction}


\begin{table*}[!t]
  \centering
  \setlength{\tabcolsep}{0.9mm} 
  \begin{tabular}{c c c c c c c c c}  
  \toprule
  \multicolumn{1}{c}{\textbf{Metric}} & \textbf{Method} & \textbf{A. Forehead} & \textbf{B. Nose} & \textbf{C. Lips} & \textbf{D. Chin} & \textbf{E. Cheek (L)} & \textbf{F. Cheek (R)} & \textbf{Entire face} \\
    \cmidrule(r){1-9}
    \multirow{5}{*}{\textbf{P2PL $\downarrow$}}
      & FSC-Net        & 2.58$\pm$1.19 & 1.90$\pm$0.62 & 2.52$\pm$0.83 & 2.74$\pm$0.92 & 2.55$\pm$0.66 & 2.77$\pm$0.81 & 2.35$\pm$0.71 \\
      & ACMT-Net & 1.12$\pm$0.34 & 1.32$\pm$0.32 & 1.56$\pm$0.49 & 2.31$\pm$0.83 & 1.21$\pm$0.31 & 1.48$\pm$0.32 & 1.29$\pm$0.26 \\
      & BPMNR       & \textbf{0.74$\pm$0.23} & \underline{0.97$\pm$0.36} & \underline{1.12$\pm$0.42} & 1.62$\pm$0.69 & \textbf{1.07$\pm$0.24} & \textbf{1.03$\pm$0.22} & \underline{0.93$\pm$0.14} \\
      & Ours (global $\text{MLP}_{\psi}$)      & \underline{0.77$\pm$0.27} & 0.87$\pm$0.33 & 1.67$\pm$0.82 & \underline{1.60$\pm$0.53} & 1.17$\pm$0.37 & 1.41$\pm$0.44 & 1.03$\pm$0.21 \\
      & Ours (separate $\text{MLP}_{\psi}$)    & 0.78$\pm$0.16 & \textbf{0.75$\pm$0.17} & \textbf{1.05$\pm$0.40} & \textbf{1.48$\pm$0.36} & \underline{1.15$\pm$0.23} & \underline{1.23$\pm$0.44} & \textbf{0.88$\pm$0.08} \\
    \cmidrule(r){1-9}
    \multirow{5}{*}{\textbf{CD} $\downarrow$}
      & FSC-Net     & 5.51$\pm$2.05 & 4.37$\pm$1.02 & 5.04$\pm$1.31 & 6.44$\pm$1.52 & 6.15$\pm$1.07 & 6.61$\pm$1.18 & 5.23$\pm$1.18 \\
      & ACMT-Net & 3.18$\pm$0.65 & 3.37$\pm$0.56 & 3.84$\pm$1.04 & 5.83$\pm$1.30 & 4.21$\pm$0.51 & 4.35$\pm$0.88 & 3.58$\pm$0.51 \\
      & BPMNR       & 2.50$\pm$0.39 & 2.88$\pm$0.58 & \underline{2.89$\pm$0.66} & 4.40$\pm$1.27 & 3.70$\pm$0.38 & \textbf{3.57$\pm$0.71} & 2.93$\pm$0.21 \\
      & Ours (global $\text{MLP}_{\psi}$)      & \underline{2.42$\pm$0.45} & \underline{2.38$\pm$0.62} & 3.76$\pm$1.53 & 4.33$\pm$0.90 & \underline{3.62$\pm$0.67} & 3.98$\pm$0.86 & \underline{2.90$\pm$0.40} \\
      & Ours (separate $\text{MLP}_{\psi}$)    & \textbf{2.38$\pm$0.25} & \textbf{2.27$\pm$0.24} & \textbf{2.61$\pm$0.72} & \textbf{4.18$\pm$0.62} & \textbf{3.61$\pm$0.33} & \underline{3.76$\pm$0.69} & \textbf{2.65$\pm$0.12} \\
    \bottomrule
  \end{tabular}
  \caption{Quantitative results compared with FSC-Net, ACMT-Net, and BPMNR, including ablation studies on postoperative facial appearance prediction accuracy, measured in millimeters (mm). Lower values indicate better performance. The best and second-best performances are shown in \textbf{bold} and \underline{underline}, respectively.}
  \label{tab:surgery_prediction}
\end{table*}

We evaluate the performance of our framework on postoperative facial appearance prediction against state-of-the-art methods. As shown in Table~\ref{tab:surgery_prediction}, our method consistently achieves the lowest P2PL and CD errors across most facial sub-regions, with marked improvements in critical areas such as the lips and chin.
Our method yields the lowest P2PL on the entire face (0.88$\pm$0.08 mm), and outperforms BPMNR notably in the lips (1.05$\pm$0.40 vs. 1.12$\pm$0.42 mm) and chin (1.48$\pm$0.36 vs. 1.62$\pm$0.69 mm). It also attains the best CD overall (2.65$\pm$0.12 mm), indicating robust surface-level accuracy across regions. These results underscore the advantage of our anatomically informed, region-specific deformation modeling.
Fig.~\ref{surgery prediction} showcases the qualitative P2PL error maps across representative cases, where our method exhibits notably fewer high-error regions and a more consistent error distribution, demonstrating superior accuracy in anatomically complex areas.
Facial appearance prediction for a single patient on an RTX 4090 requires around 150 seconds in total—shape latent codes fitting ($\sim$70 s), surgery latent codes fitting ($\sim$50 s), and mesh generation ($\sim$30 s)—with a peak memory footprint of around 1.25 GB.

\subsection{Ablation Studies}
We further perform ablation studies on the internal architecture components of both the shape and surgery module, to assess their contributions to the overall performance.



For the shape module, we compare two configurations: one employing three region-specific SDF decoders (denoted as multi decoders), and the other using a single shared decoder (single decoder). This evaluates whether anatomical specialization improves geometric reconstruction.
For the surgery module, we assess the role of the dimensionality reduction network $\text{MLP}_{\psi}$. Specifically, we compare a shared global $\text{MLP}_{\psi}$ that processes all region-specific latent codes jointly with separate $\text{MLP}_{\psi}$ networks, each handling its corresponding region. This investigates whether region-specific modeling enhances deformation prediction.

The ablation results in Table~\ref{tab:shape_surgery_joint}, and Table~\ref{tab:surgery_prediction} show that region-specific designs consistently improve the performance of both shape reconstruction and surgery deformation modeling. For the shape module, using separate SDF decoders better captures anatomical variations. In the surgery module, adopting region-specific $\text{MLP}_{\psi}$ leads to more accurate deformation modeling across facial components. As shown in Fig.~\ref{eval of shape and module} and Fig.~\ref{surgery prediction}, these configurations produce smoother surfaces and more precise geometry, particularly in complex regions such as the lips, chin, and temporomandibular joint. These results highlight the benefit of anatomical decomposition in enhancing model fidelity.

\section{Conclusion}
In this work, we presented Neural Implicit Craniofacial Model (NICE) for accurate postoperative facial appearance prediction in orthognathic surgery. By leveraging region-specific implicit SDF decoders and shape latent representations, our shape module enables high-fidelity reconstruction of both facial soft tissue and bony structures. The surgery module further introduces anatomically guided, region-specific deformation decoders driven by a unified surgery latent code, effectively capturing complex nonlinear biomechanical responses of facial tissues to skeletal movements. Comprehensive experiments on clinical data demonstrate that our method outperforms existing state-of-the-art approaches in both geometric accuracy and anatomical consistency, particularly in critical regions such as the lips and chin.
Future work will enhance latent-space interpretability and control by adding landmark-guided constraints. Although the current latent codes effectively encode geometry and deformation, they lack explicit semantic correspondence. We aim to establish a more structured latent space that enables precise, anatomically meaningful control over surgical outcomes.


\section*{Acknowledgments}
 This work was supported by the National Natural Science
 Foundation of China under Grant no. 62571328. The experiments of this work were supported by the core facility Platform of Electronics, SIST, ShanghaiTech University.

\bibliography{aaai2026}


\end{document}